\documentclass[letterpaper, 10 pt, conference]{ieeeconf}  

\IEEEoverridecommandlockouts                              

\usepackage{graphics} 
\usepackage{epsfig} 
\usepackage{mathptmx} 
\usepackage{times} 
\usepackage{amsmath} 
\usepackage{amssymb}  
\usepackage{float}
\usepackage{orcidlink}
\usepackage{booktabs}
\usepackage{multirow}
\usepackage{caption}
\usepackage{color}
\usepackage{colortbl}
\usepackage{booktabs}
\usepackage{multirow}
\usepackage{xcolor}
\usepackage{caption}
\usepackage{gensymb}
\usepackage{url}
\usepackage{amsmath} 
\usepackage{amssymb}  
\usepackage{subfigure}
\usepackage{array}
\usepackage{soul}
\usepackage{booktabs}
\usepackage{tcolorbox}
\usepackage{color,xcolor}
\usepackage{multirow}

\usepackage{float}
\usepackage{cite}
\usepackage{algpseudocode}
\usepackage[export]{adjustbox}
\usepackage{balance}
\usepackage{rotating}
\usepackage{graphicx}
\usepackage{hyperref}

\definecolor{mygreen}{RGB}{41,200,51}
\definecolor{myred}{RGB}{160,30,30}
\definecolor{myblue}{RGB}{255,0,0}

\title{\LARGE \bf
MVCTrack: Boosting 3D Point Cloud Tracking via Multimodal-Guided Virtual Cues
}

\author{Zhaofeng Hu$^{1}$$^\dagger$, Sifan Zhou$^{2}$$^\dagger$*, Zhihang Yuan$^{3}$, Dawei Yang$^{3}$, Shibo Zhao$^{4}$, Ci-jyun Liang$^{1}$*
\thanks{$^{1}$ Stony Brook University. $^{2}$ Southeast University. $^{3}$ Houmo AI. $^{4}$ Carnegie Mellon University. $^\dagger$ Authors with equal contribution.}
\thanks{*Corresponding author: sifanjay@gmail.com, ci-jyun.liang@stonybrook.edu.}
}

\begin{document}

\maketitle
\thispagestyle{empty}
\pagestyle{empty}

\begin{abstract}

3D single object tracking is essential in autonomous driving and robotics. Existing methods often struggle with sparse and incomplete point cloud scenarios. To address these limitations, we propose a Multimodal-guided Virtual Cues Projection (MVCP) scheme that generates virtual cues to enrich sparse point clouds. Additionally, we introduce an enhanced tracker MVCTrack based on the generated virtual cues. Specifically, the MVCP scheme seamlessly integrates RGB sensors into LiDAR-based systems, leveraging a set of 2D detections to create dense 3D virtual cues that significantly improve the sparsity of point clouds. These virtual cues can naturally integrate with existing LiDAR-based 3D trackers, yielding substantial performance gains. Extensive experiments demonstrate that our method achieves competitive performance on the NuScenes dataset. Code is available at \href{https://github.com/StiphyJay/MVCTrack}{\raisebox{-0.05\height} \ \textbf{\textcolor{red}{{code}}}} and \href{https://youtu.be/c-OPJ0PvvbA}{\raisebox{-0.05\height} \ \textbf{\textcolor{red}{{video}}}}.

\end{abstract}
%


\section{INTRODUCTION}
Recently, LiDAR-based 3D single object tracking (3D SOT) has garnered attention due to its wide applications in autonomous driving\cite{probabilistic,ptt,zhou2023fastpillars} and mobile robots\cite{eagermot,realtime}. Compared to RGB cameras, LiDAR can easily gauge spatial distances, relationships and shapes of objects by collecting laser measurement signals to represent 3d models and maps of environments, exhibiting robustness against visual degraded. Those advantages makes LiDAR particularly appealing for tracking tasks which the scenarios and illumination always change rapidly. Most existing 3D SOT methods \cite{cui2019point,p2b,3dsiamrpn, ptt, ptt-journal,lttr,bat,v2b} inherit the 2D visual tracking pipeline, which leverage siamese networks~\cite{siamfc,siamrpn,siamla} for feature extraction and geometry matching between the template and search region. Despite being effective, the inherent sparsity and low resolution of LiDAR still lead to unsatisfactory performance, especially in distant-range scenarios and small size objects (e.g., pedestrian, cyclist). Notably, compact and cheap RGB cameras can provide dense semantic and texture features, effectively mitigating the inherent sparsity of point clouds and assist trackers in distinguishing targets from disturbances. Therefore, a natural and direct approach is to leverage the complementary multi-modal information to boost the performance of 3D SOT tasks.  
\begin{figure}[t]
    \centering
    \includegraphics[width=\linewidth]{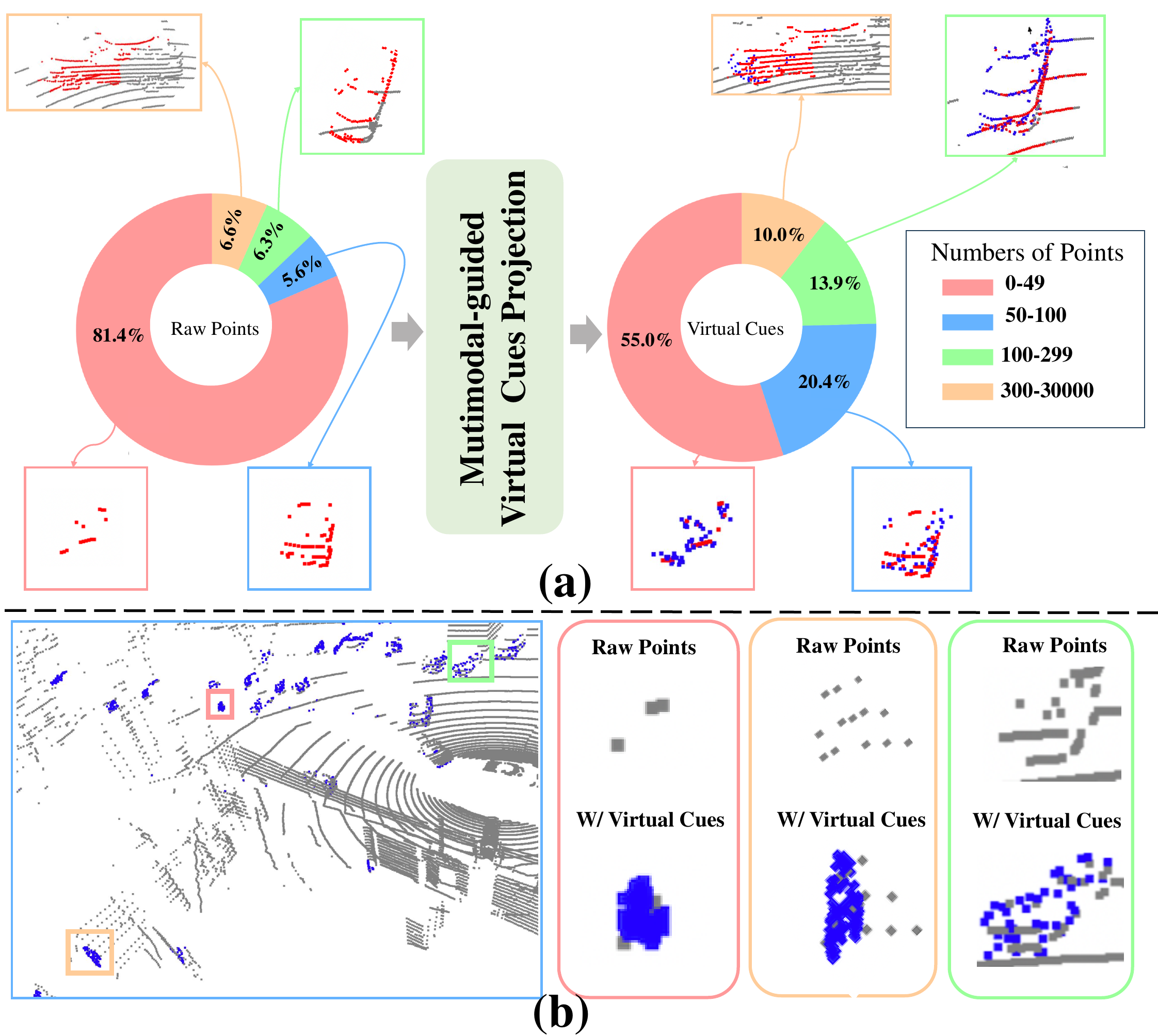}
    \caption{\textbf{(a)} Left: statistics of the number of points on nuScenes's car. Right: the number of points on nuScenes's car with multimodal-guided virtual cues. Raw points is \textcolor{red}{red}, the virtual cues is \textcolor{blue}{blue}. \textbf{(b)} Virtual cues generated using MVCP scheme. Blue Square box: Virtual cues of a certain frame. The raw points are marked in \textcolor{gray}{gray}, and the virtual cues are marked in \textcolor{blue}{blue}.} 
    \label{fig:points_distribution}
    \vspace{-8mm}
\end{figure}

F-Siamese~\cite{F-siamese} is the first multi-modal SOT tracker. It initially employs a 2D tracker to estimate the 2D bounding box of the target, which is then projected into the 3D viewing frustum. Subsequently, it introduces a frustum-based dual Siamese network that effectively reduces redundant point cloud area by combining 2D region proposals with the 3D search space, thereby improving tracking accuracy. However, F-Siamese overlooks the feature interactions between the different modalities, resulting in the loss of critical cues from dense images. Thus, its performance lags much behind LiDAR-Only methods. After that, MMF-Track \cite{mmftrack} further enhances the semantic association between point clouds and images through multi-level feature interaction and fusion, significantly improving the model's performance in sparse and occluded scenarios. However, its complex feature interaction module greatly limits the inference speed, which is crucial tracking tasks with real-time requirements.


Unlike the aforementioned methods that introduce additional visual branches to generate frustums or dense RGB features for semantic-texture fusion, we argue that directly leveraging virtual cues from the RGB image modality can effectively enhance the performance\cite{mvp} without incurring significant additional computational overhead. Notably, many existing object detection and segmentation methods~\cite{probabilistic,yolov6,yolov7} are lightweight and highly efficient, making them well-suited for use on resource-limited edge devices. These methods can maintain high perception performance while introducing minimal latency, making them particularly suitable for latency-sensitive tracking tasks. Building on this intuition, we propose a novel \textbf{Multimodal-guided Virtual Cues Projection (MVCP)} mechanism to enhance the density and completeness of point clouds, thereby mitigating the impact of sparse point cloud on accuracy and facilitating a better understanding of the surrounding environment. As shown in Fig\ref{fig:points_distribution}, we analyze the number of points in the nuScenes~\cite{nuScenes} dataset. It can be found that 81\% of objects contain fewer than 50 points, while only 7\% of objects have more than 300 points. This significant sparsity issue severely limits the tracking performance, as the few discriminative points on the target make it difficult to distinguish from the background. However, after applying our proposed Multimodal-guided Virtual Cues Projection (MVCP) mechanism, a significant change in the distribution of point clouds in the original dataset is observed. The proportion of objects with fewer than 50 points decreases from 81\% to 55\%, while the proportion of targets with more than 50 points rises from 19\% to 45\%. 

Particularly, in our Multimodal-guided Virtual Cues Projection (MVCP) mechanism, a lightweight 2D object segmentor~\cite{probabilistic} is employed to crop the original point cloud into instance-wise frustums. Subsequently, dense 3D virtual cues are generated near these foreground points by lifting 2D pixels into 3D space, with depth completion used in the image space to infer the depth of these virtual cues. Finally, the MVCP combines the virtual cues with the original LiDAR measurements as input to a 3D SOT tracker~\cite{p2p}. Based on MVCP scheme, we construct our \textbf{MVCTrack} framework, which offers three key advantages: \textbf{(1) Lightweight 2D object detection} The 2D object detectors employed are well-optimized~\cite{probabilistic, yolov7}, ensuring that their integration does not incur significant computational overhead while effectively leveraging dense visual features. \textbf{(2) Balanced point density distribution}: The virtual cues balance the density distribution of points across different distances, reducing the density imbalance between near and far objects. \textbf{(3) Plug-and-play module}: Our projection mechanism serves as a plug-and-play module that can be integrated with any existing or new advanced 2D or 3D detectors yielding substantial benefits with minimal effort—achieving significant gains. We evaluate our MVCTrack on the large-scale nuScenes datasets~\cite{nuScenes}. Extensive experimental results demonstrate that our approach achieves competitive performance on the nuScenes dataset, significantly surpassing existing multi-modal 3D trackers. Our main contributions are summarized as follows:

\begin{itemize}

\item \textbf{Multimodal-guided Virtual Cues Projection (MVCP).} A novel plug-in scheme for generating virtual cues that leverages dense semantics from the images to compensate for the sparsity of point clouds.

\item \textbf{MVCTrack.} An enhanced 3D single object tracking network that integrates virtual cues, enabling end-to-end training to improve overall tracking performance, particularly for small and distant objects.

\item Our approach achieves competitive performance on the challenging large-scale nuScenes datasets. Extensive ablation studies demonstrate the effectiveness and generalization of the proposed methods. 

\end{itemize}
\vspace{-2mm}
\section{RELATED WORK}
\noindent\textbf{LiDAR-based 3D Single Object Tracking.} Leveraging the insensitivity of LiDAR to illumination changes and its ability to capture accurate distance information, numerous works on LiDAR-based 3D single object tracking (3D SOT) have emerged. SC3D~\cite{sc3d}, as a pioneering work in 3D SOT, employs a Kalman filter to generate a set of candidate 3D bounding boxes and selects the one most similar to the template target as the predicted result. However, due to the time-consuming candidate generation process, SC3D is not an end-to-end framework and struggles to achieve real-time performance. To address these problems, P2B~\cite{p2b} introduced a target-specific feature matching framework and utilized VoteNet\cite{votenet} to estimate the target center. Similarly, 3D-SiamRPN~\cite{3dsiamrpn} implemented a region proposal network for object tracking. Building upon this foundation, BAT\cite{bat} incorporated box-aware information to enhance similarity features. PTT~\cite{ptt, ptt-journal} proposes the Point-Track-Transformer module to assign weights to crucial point cloud features. STNet\cite{stnet} developed an iterative coarse-to-fine correlation network for robust correlation learning. GLT-T\cite{glt} introduced a global-local Transformer voting scheme to generate higher-quality 3D proposals. A series of follow-up methods \cite{pttr,stnet,cmt,dmt,glt,osp2b,tat,synctrack,xu2024pillartrack} have also adopted appearance matching frameworks. In contrast to the appearance matching framework, M$^2$Track~\cite{m2track} introduced a motion-centric framework that utilizes motion information rather than appearance for 3D SOT, achieving impressive results. Furthermore, M$^2$Track++~\cite{m2track++} explored the performance with a semi-supervised setting. However, those methods are still limited by the sparsity of point cloud in single modal.


\noindent\textbf{Multi-Model 3D Single Object Tracking.} Multi-Model 3D SOT leverage the complementary nature of diverse sensor data, such as LiDAR and RGB, to achieve more robust and accurate tracking. F-Siamese\cite{F-siamese} introduced a frustum-based dual Siamese network that extends 2D region proposals from RGB images into 3D space, effectively reducing redundant 3D search areas. By performing accuracy validation within the 3D frustum, F-Siamese maintained a certain success rate even in cases of sparse point clouds or target occlusion; however, it lacks feature-level alignment and exhibits relatively slow tracking speed. Subsequently, MMF-Track \cite{mmftrack} proposed a multimodal, multi-level fusion approach that aligns RGB images and point clouds in 3D space through a spatial alignment module, facilitating feature-level interaction between geometric and texture information. Although the multi-layer feature interaction in MMF-Track enables robust tracking performance in complex environments, its intricate feature interaction process significantly constrains inference speed, particularly on onboard robotic platforms.

\begin{figure*}[t]
    \centering
    \vspace{2mm}
    \includegraphics[width=0.9\linewidth]{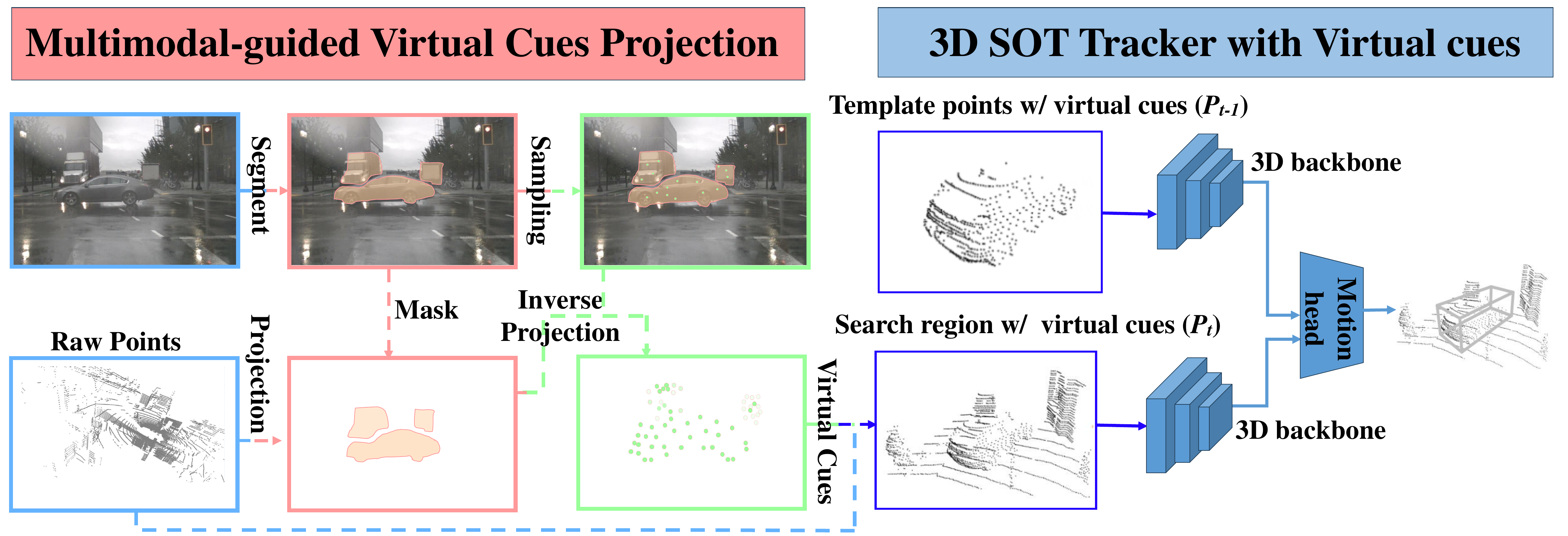}
    \caption{MVCTrack Framework. Firstly, Multimodal-guided virtual cues projection, this step will generate virtual cues based on the segmentation masks and raw points. Secondly, 3D single object tracking with the integration of raw points and virtual cues.} 
    \label{fig:framework}
    \vspace{-5mm}
\end{figure*}

\section{PRELIMINARY}
\noindent\textbf{3D Single Object Tracking.} In 3D SOT, given the point cloud input \( P = \{(x_i, y_i, z_i)\} \) at time $t$, where \( (x_i, y_i, z_i) \) denotes the 3D coordinates of a point. The 3D SOT task aims to predict the 3D bounding box of the tracked object $\mathbf{b}_t = (u, v, o, w, l, h, \theta)$, where $(u, v, o)$ represent the 3D center location, $w, l, h$ denote the width, length, and height of the object, and $\theta$ represents the rotation around the z-axis. The primary goal is to maintain a consistent tracking of the object across sequential frames by exploiting the spatio-temporal information provided by the point clouds.

\noindent\textbf{2D-3D Correspondence.} We establish a correspondence between 2D image pixels and 3D point cloud data. This transformation process involves a series of homogeneous transformations, accounting for sensor alignment and motion compensation. Specifically, we apply rotation and translation to align LiDAR points with the RGB image frame. Here, $X_{\text{car} \leftarrow \text{lidar}}$ represents the transformation from the LiDAR coordinate frame to the vehicle’s reference frame, while $X_{t_1 \leftarrow t_2}$ accounts for motion compensation between timestamps $t_1$ and $t_2$. The transformation $X_{\text{rgb} \leftarrow \text{car}}$ maps points from the vehicle’s reference frame to the RGB camera frame. Finally, the RGB camera's intrinsic matrix $I_{\text{rgb}}$ is used to project the 3D points onto the 2D image plane.


\vspace{-2mm}
\begin{gather}
    \mathit{X}_{t_1 \leftarrow t_2}^{\text{rgb} \leftarrow \text{lidar}} = \mathit{X}_{\text{rgb} \leftarrow \text{car}} \cdot \mathit{X}_{t_1 \leftarrow t_2} \cdot \mathit{X}_{\text{car} \leftarrow \text{lidar}} \label{Eq:transformation} \\
    \mathbf{P}_{\text{rgb}} = I_{\text{rgb}} \cdot \mathit{X}_{t_1 \leftarrow t_2}^{\text{rgb} \leftarrow \text{lidar}} \cdot \mathbf{P}_{\text{lidar}} \label{eq:attention_prefill}
\end{gather}
\vspace{-5mm}

\section{METHOD}
\subsection{Overview}

Given an initial bounding box $B_0 = (x_0, y_0, z_0, w, h, l, \theta_0) \in \mathbb{R}^7$, where $(x_0, y_0, z_0)$ denotes the object center, $(w, h, l)$ represent the object’s size, and $\theta_0$ indicates its orientation. Throughout tracking, we predict the target state $B_f = (x_f, y_f, z_f, \theta_f) \in \mathbb{R}^4$ for each frame $f$, because the size $(w, h, l)$ remains constant across frames. Our method leverages multi-modal virtual cues $v_i = (x, y, z)$, where $(x, y, z)$ specifies the 3D coordinates obtained from MVCP scheme. For each 2D object detection $b_j$ with an associated instance mask $m_j$, we generate a fixed number $\tau$ as virtual cues to compensate sparsity of the point cloud. 

\subsection{Multimodal-guided Virtual Cues Projection}

\subsubsection{Virtual Cues Sampling}

The process of virtual cues sampling bagins with identifying potential areas around the tracked object where additional information may enhance tracking performance. Given the inherent sparsity of LiDAR point clouds, particularly around small or distant objects, we aim to densely populate these regions with additional points not originally captured by the LiDAR sensor. To achieve this, we leverage 2D image data, which often provides dense and detailed object boundaries, to inform the placement of virtual cues in 3D space.

Typically, we obtain a set of 2D object segmentation masks $\mathbf{B}_{obj} = \{b_1, b_2, \dots, b_n\}$ from an 2D segmentor  at every keyframe. Each object is represented by a bounding box $b_j = (u_j, v_j, w_j, h_j)$ and an associated instance mask $m_j$, which segments the object’s pixels in the 2D image. Using this mask, we generate a fixed number $\tau$ of virtual cues $\mathbf{v}_i = (x_i, y_i, z_i)$ uniformly across the detected object’s area in the image, where $(x_i, y_i, z_i)$ is the 3D location of generated virtual cues. 
\vspace{-2mm}
\begin{equation}
    z_i = \textbf{NearestNeighborDepth}(x_i, y_i) \label{Eq:depth_estimation}
\end{equation} 
\vspace{-2mm}
\begin{equation}
    v_i = X_{\text{lidar} \leftarrow \text{rgb}}^{-1} \cdot X_{\text{rgb} \leftarrow \text{cam}}^{-1} \cdot I_{\text{rgb}}^{-1} \cdot 
    \begin{bmatrix} 
    x_i \\ 
    y_i \\ 
    1 
    \end{bmatrix} 
    \cdot z_i, \quad i = 1,\dots, \tau
    \label{Eq:2D_to_3D}
\end{equation}

\vspace{-0mm}

where $\tau$ is the sampled number of virtual cues per object.

\subsubsection{Virtual Cues Projection}
To project the virtual cues from 2D into 3D space, we use the LiDAR point cloud to infer depth. Each point $\mathbf{p}_i = (x_i, y_i, z_i)$ from the LiDAR point cloud is first projected onto the 2D image plane, and for each virtual point $\mathbf{v}_i = (x_i, y_i, z_i)$, we assign the depth value $d_i$ of its nearest neighbor LiDAR point in the projected 2D space:
\vspace{-2mm}
\begin{equation}
d_i = \arg \min_{\mathbf{p}_i} \left\| \mathbf{v}_i - \mathbf{p}_i^{\text{2D}} \right\|
\vspace{-2mm}
\end{equation}

The virtual cues are unprojected back into 3D space, ensuring that each virtual point $\mathbf{v}_i$ has a full 3D coordinate $(x_i, y_i, z_i)$:
\vspace{-3mm}
\begin{equation}
\mathbf{v}_i = \left( \frac{x_i}{z_i}, \frac{y_i}{z_i}, z_i\right)
\vspace{-2mm}
\end{equation}

As a result of the above process, the denser point cloud is generated, which effectively enhances the semantic information of discriminative points in sparse or incomplete LiDAR data. More advanced depth association methods could yield virtual cues with finer-grained geometric information. We remain this issue for future work.

\begin{table*}[t]
\centering
\caption{Comparisons with state-of-the-art methods on nuScenes dataset~\cite{nuScenes}. L means LiDAR-Only methods. LC denotes LiDAR-Camera methods. M and S are motion-based and similarity-based paradigms. \textit{Success} / \textit{Precision} are used for evaluation. \textbf{Bold} and \underline{underline} denote the best result and the second-best one, respectively. $\dagger$ means our reimplementation based on official code.}
    \resizebox{1.0\textwidth}{!}{
    \normalsize
    \begin{tabular}{c|c|c|ccccc|c}
          \toprule
          Tracker & Publish & Modality & Car [64,159] & Pedestrian [33,227] & Truck [13,587] & Trailer [3,352] & Bus [2,953] & Mean [117,278] \\
          \midrule
          SC3D~\cite{sc3d} & CVPR'19 & \multirow{12}{*}{L+S} & 22.31 / 21.93 & 11.29 / 12.65 & 30.67 / 27.73 & 35.28 / 28.12 & 29.35 / 24.08 & 20.70 / 20.20 \\
          P2B~\cite{p2b} & CVPR'20 & & 38.81 / 43.18 & 28.39 / 52.24 & 42.95 / 41.59 & 48.96 / 40.05 & 32.95 / 27.41 & 36.48 / 45.08 \\
          PTT~\cite{ptt-journal} & IROS'21 & & 41.22 / 45.26 & 19.33 / 32.03 & 50.23 / 48.56 & 51.70 / 46.50 & 39.40 / 36.70 & 36.33 / 41.72 \\
          BAT~\cite{bat} & ICCV'21 & & 40.73 / 43.29 & 28.83 / 53.32 & 45.34 / 42.58 & 52.59 / 44.89 & 35.44 / 28.01 & 38.10 / 45.71 \\
          V2B~\cite{v2b} & NIPS'21 & & 54.40 / 59.70 & 30.10 / 55.40 & 53.70 / 54.50 & 54.90 / 51.44 & - / - & - / - \\
          PTTR~\cite{pttr} & CVPR'22 & & 51.89 / 58.61 & 29.90 / 45.09 & 45.30 / 44.74 & 45.87 / 38.36 & 43.14 / 37.74 & 44.50 / 52.07 \\
          M$^2$Track~\cite{m2track} & CVPR'22 & & 55.85 / 65.09 & 32.10 / 60.92 & 57.36 / 59.54 & 57.61 / 58.26 & 51.39 / 51.44 & 49.23 / 62.73 \\
          SMAT~\cite{smat} & RAL'22 & & 43.51 / 49.04 & 32.27 / 60.28 & - / - & - / - & 39.42 / 34.32 & - / - \\
          STTracker~\cite{stttracker} & RAL'23 & & 56.11 / 69.07 & 37.58 / 68.36 & 54.29 / 60.71 & 36.31 / 36.07 & 48.13 / 55.48 & 49.88 / 66.61 \\
          GLT-T~\cite{glt} & AAAI'23 & & 48.52 / 54.29 & 31.74 / 56.49 & 52.74 / 51.43 & 57.60 / 52.01 & 44.55 / 40.69 & 44.42 / 54.33 \\
          MoCUT~\cite{cutrack} & ICLR'24 & & 57.32 / 66.01 & 33.47 / 63.12 & 61.75 / 64.38 & 60.90 / 61.84 & 57.39 / 56.07 & 51.19 / 64.63 \\
          FlowTrack~\cite{flowtrck} & IROS'24 & & 60.29 / 71.07 & 37.60 / 67.64 & - / - & 55.39 / 62.70 & - / - & - / - \\ 
          \midrule
          MMFTrack~\cite{mmftrack} & TIV'23 & LC+S & 50.73 / 58.51 & 32.80 / 66.25 & - / - & - / - & 52.73 / 52.78 & - / - \\
          \midrule
          Baseline~\cite{p2p}$^\dagger$ & - & L+M & \underline{64.61} / \underline{71.98} & \underline{45.64} / \underline{74.62} & \underline{64.42} / \underline{65.37} & \underline{70.23} / \underline{66.08} & \underline{58.54} / \underline{56.13} & \underline{59.22} / \underline{71.19} \\
          \midrule
          \textbf{MVCTrack} & \multirow{2}{*}{\textbf{Ours}} & \multirow{2}{*}{LC+M} & \textbf{66.76} / \textbf{73.76} & \textbf{47.34} / \textbf{76.64} & \textbf{66.21} / \textbf{66.33} & \textbf{72.73} / \textbf{69.80} & \textbf{60.20} / \textbf{58.88} & \textbf{61.20} / \textbf{73.22} \\
          Improvement & & & \textcolor{mygreen}{$\uparrow$2.15} / \textcolor{mygreen}{$\uparrow$1.78} & \textcolor{mygreen}{$\uparrow$1.70} / \textcolor{mygreen}{$\uparrow$2.02} & \textcolor{mygreen}{$\uparrow$1.79} / \textcolor{mygreen}{$\uparrow$0.96} & \textcolor{mygreen}{$\uparrow$2.50} / \textcolor{mygreen}{$\uparrow$3.72} & \textcolor{mygreen}{$\uparrow$1.66} / \textcolor{mygreen}{$\uparrow$2.75} & \textcolor{mygreen}{$\uparrow$1.98} / \textcolor{mygreen}{$\uparrow$2.03} \\
          \bottomrule
    \end{tabular}%
}%

\label{table3}
\vspace{-2mm}
\end{table*}

\subsection{MVCTrack with Virtual Cues}

To validate the effectiveness of generated virtual cues for 3D SOT tasks, we integrate those virtual cues to enhance the detail of object boundaries by concatenating with the raw points. Specifically, the raw LiDAR point cloud $P = \{(x_i, y_i, z_i)\}$ is combined with the virtual point cloud $\mathbf{V} = \{(x_i, y_i, z_i)\}$, forming an augmented point cloud $\mathbf{P}^{\text{aug}} = P \cup \mathbf{V}$. Subsequently, the augmented point cloud are as follows:
\begin{equation}
\mathbf{P}^{\text{xyz}} = \left\{ (x_i, y_i, z_i) \mid (x_i, y_i, z_i) \in P \cup \mathbf{V} \right\}
\vspace{-2mm}
\end{equation}
where $\quad \mathbf{P}^{\text{xyz}} \in \mathbb{R}^{n \times 3}$. The augmented point cloud $\mathbf{P}$ contains rich geometric and semantic information that effectively reduces noise interference and enhances the representation of target objects. The overall framework of MVCTrack is shown in Fig \ref{fig:framework}, where the augmented point cloud is input into a 3D convolutional network for encoding, capturing both spatial and semantic features of the objects. After the convolutional operations, the network transforms the extracted 3D feature map into a BEV feature map, which contains global spatial information about the object in the plane. Guided by virtual cues, the model can more clearly identify the central position, contour, and orientation angle $\theta$ of the target object in the BEV feature map, thereby improving the accuracy of the 3D bounding box. Finally, based on the BEV feature map, the model regresses the final 3D bounding box $B_f = (x_f, y_f, z_f, \theta_f)$ for the target object and achieves continues tracking in subsequent frames. Our MVCTrack not only compensates for deficiencies in sensor data but also significantly improve the tracking accuracy through explicitly multi-modal information fusion. More details about the tracking network can refer to ~\cite{p2p}.

\section{EXPERIMENTS}
\subsection{Dataset, Metrics and Implementation Details.}
We follow the common setup~\cite{p2b,m2track} and conduct experiments on the large-scale nuScenes~\cite{nuScenes} dataset. Notably, due to the limited sample size of the KITTI dataset (only 19 training, 2 validation sequences~\cite{ptt,glt,mbptrack}) makes it challenging to adequately evaluate the methods. In contrast, the nuScenes dataset comprises 700 training and 150 validation sequences, allowing for a more comprehensive assessment. The evaluation metrics is followed the common setup~\cite{ptt,glt,mbptrack} to report \textit{Success} and \textit{Precision} based on one pass evaluation (OPE)~\cite{otb2013,kristan2016novel}. We follow existing methods~\cite{ptt,p2p} and set the extended range as 
$[(-4.8, 4.8), (-4.8, 4.8), (-1.5, 1.5)]$ for cars and 
$[(-1.92, 1.92), (-1.92, 1.92), (-1.5, 1.5)]$ for humans along the $(x, y, z)$ axes. The extended range estimates the target's potential position in the next frame. The backbone of MVCTrack is designed with 4 sparse convolution blocks, each featuring {16, 32, 64, 128} channels, as the Shared backbone for efficient feature extraction. The tracking head is designed with 2 fully connected layers followed by a sigmoid activation function, which outputs the predicted center position and orientation. And we adopt a single regression loss to minimize the distance between predicted and ground truth center positions, along with an angular difference penalty for orientation. The entire MVCTrack network can be trained end-to-end. We train the model on two NVIDIA GTX 4090 GPUs for a total of 20 epochs using the AdamW optimizer, with a learning rate set to 1e-4 and a batch size of 256. The weight decay parameter is fixed at 1e-5. During training, we apply common data augmentation strategies such as random flipping, random rotation, and random translation. The loss function used is a single regression loss to refine the predicted outputs effectively.  Our experiment video is available at \href{https://youtu.be/c-OPJ0PvvbA}{\raisebox{-0.05\height} \ \textbf{\textcolor{red}{{video}}}}.

\subsection{Quantitative, Qualitative, and Ablation Study}
\noindent\textbf{Extensive results on nuScenes.} 
We evaluate our MVCTrack on the large-scale nuScenes~\cite{nuScenes} dataset, known for its challenging and sparse point clouds. As shown in Tab.~\ref{table3}, MVCTrack achieves significant improvements over prior methods and baseline, outperforming M$^2$Track, which also uses motion cues, by 10.91\%/8.67\% and 15.23\%/15.75\% in Car and Pedestrian. Compared with the latest multi-modal tracking methods MMF-Track\cite{mmftrack}, MVCTrack outperforms 16.03\%/15.25\% and 14.53\%/10.39\% in the categories of Cars and Pedestrians. And MVCTrack also outperforms the baseline~\cite{p2p} without using virtual clues by 1.98\%/2.03\% in the mean accuracy, demonstrating the effectiveness of our multimodal-guided virtual cues Projection scheme.


\begin{figure*}[t]  
    \centering

    \includegraphics[width=\textwidth, trim=55 000 280 00, clip]{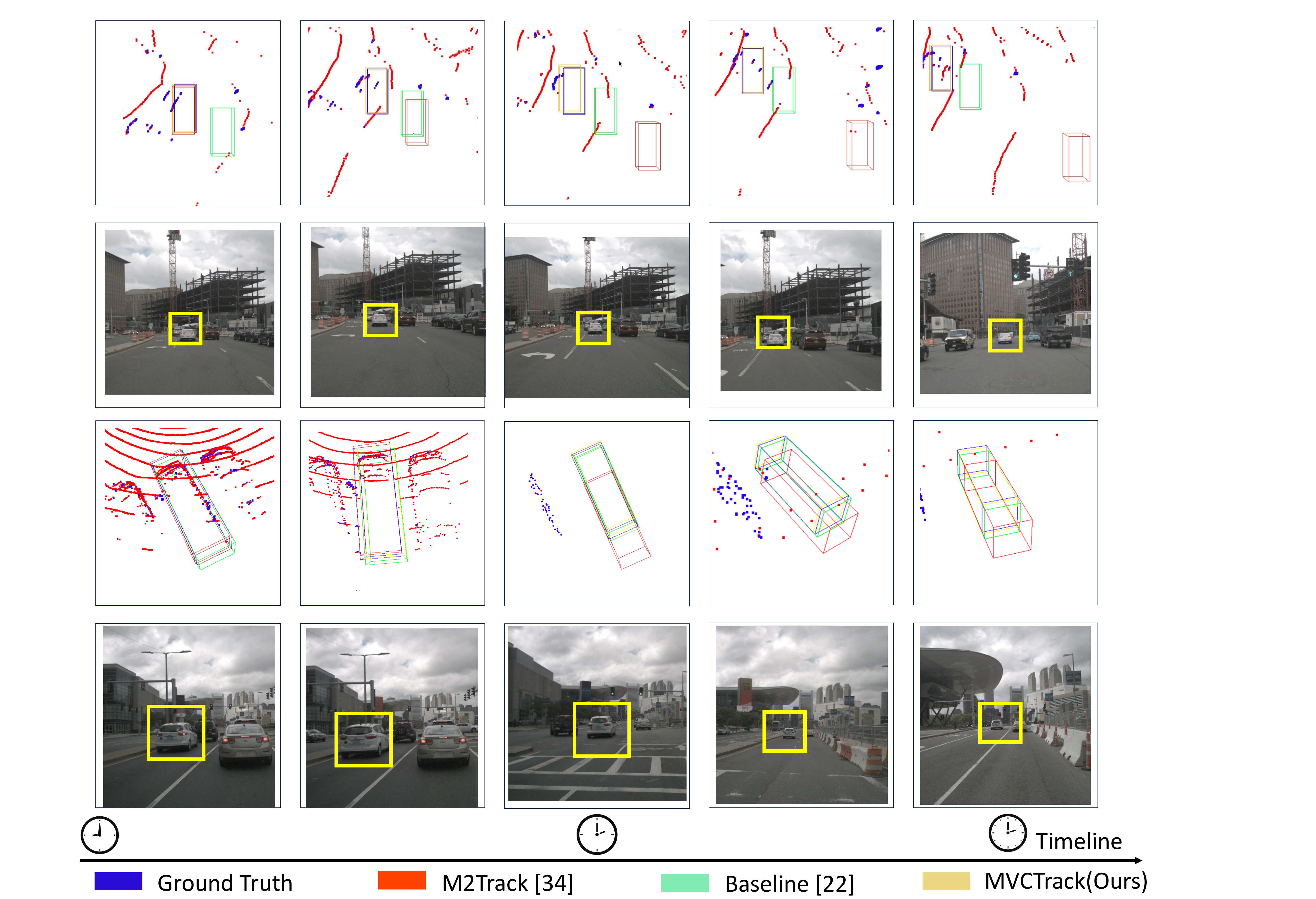}
    \caption{Visualization results on nuScenes dataset. We compare our MVCTrack with M$^2$Track~\cite{m2track} and baseline model~\cite{p2p}.}  
    \label{fig:vis}  
    \vspace{-5mm}
\end{figure*}



\noindent\textbf{Ablation study of sampling strategy.} MVCP is crucial for improving tracking accuracy in sparse scenarios. Therefore, we ablation the impact of different sampling strategies. \textbf{Strategy 1}: The sampling number of virtual points are generated based on the object's distance from the sensor. Distant objects receive fewer points (with a minimum set to $\mathbf{\lambda}$), while closer objects receive more points (with a maximum set to $\mathbf{2\lambda}$). \textbf{Strategy 2}: Similar to Strategy 2, but with the converse trend: sparse objects at a distance receive more points (with a minimum set to $\mathbf{\lambda}$), while dense objects nearby receive fewer points (with a maximum set to $\mathbf{2\lambda}$). \textbf{Strategy 3}: A fixed number of virtual points $\mathbf{\lambda}$ is uniformly generated for all objects. As shown in Tab. \ref{tab:diff_strategy}, Strategy 3 consistently exhibits the best performance, which we attribute to the fixed sampling number of virtual points being more similar with the real-world distribution of point clouds across varying distances.

\begin{table}[!h]
    \caption{Comparison for different sampling strategies.}
    \centering
    \setlength{\tabcolsep}{2.5pt} 
    \renewcommand{\arraystretch}{1.1} 
    \begin{tabular}{p{0.4\linewidth}<{\centering}p{0.2\linewidth}<{\centering}p{0.2\linewidth}<{\centering}} %
        \hline
        \multirow{2}{*}{Strategy} & \multicolumn{2}{c}{Category} \\
        \cmidrule(lr){2-3}
        & Car & Pedestrian \\
        \hline
        Strategy 1 & 64.53 / 72.92 & 45.59 / 74.64 \\
        \hline
        Strategy 2 & 65.59 / 73.38  & 45.38 / 74.71 \\
        \hline
        Strategy 3 & \textbf{66.76} / \textbf{73.76} & \textbf{47.34} / \textbf{76.64} \\
        \hline
    \end{tabular}
    \label{tab:diff_strategy}
    \vspace{-5mm}
\end{table}

\noindent\textbf{The effectiveness on small objects.} In Tab~\ref{tab:performance_comparison}, we perform the accuracy on small objects such as pedestrians, bicycle. Consistent with prior 
studies\cite{m2track,mmftrack,flowtrck}, small objects typically exhibit reduced point cloud representation, making tracking particularly challenging. However, our MVCTrack shows significant performance gains in these categories. By employing MVCP scheme, our MVCTrack effectively increases the point cloud density for small objects, leading to a notable accuracy improvement. This shows the effectiveness of our approach in handling small objects.

\begin{table}[!h]
    \caption{Comparison on small objects}
    \centering
    \begin{tabular}{p{0.2\linewidth}<{\centering}p{0.2\linewidth}<{\centering}p{0.2\linewidth}<{\centering}p{0.2\linewidth}<{\centering}} %
        \toprule
        \multirow{2}{*}{Strategy/Method} & \multicolumn{2}{c}{Category} \\
        \cmidrule(lr){2-3}
        & Pedestrian & Bicycle\\
        \hline
        M$^2$Track~\cite{m2track} & 32.10 / 60.92 & 36.32 / 67.50\\
        MMFTrack~\cite{mmftrack} & 32.80 / 66.25 & 37.53 / 68.59 \\
        \hline
        Ours & \textbf{47.34} / \textbf{76.64} & \textbf{52.16} / \textbf{77.21}  \\
        Improvement& \textcolor{mygreen}{$\uparrow$14.53} / \textcolor{mygreen}{$\uparrow$9.73} & \textcolor{mygreen}{$\uparrow$14.53} / \textcolor{mygreen}{$\uparrow$9.73} \\
        \hline
    \end{tabular}
    \vspace{-5mm}
    \label{tab:performance_comparison}
\end{table}

\begin{table}[!h]
    \caption{Comparison in long-distance scenarios.}
    \centering
    \setlength{\tabcolsep}{3pt} 
    \renewcommand{\arraystretch}{1.2} 
    \begin{tabular}
    {p{0.12\linewidth}
    <{\centering}p{0.1\linewidth}
    <{\centering}p{0.32\linewidth}
    <{\centering}p{0.32\linewidth}<{\centering}p{0.2\linewidth}<{\centering}p{0.2\linewidth}} %
        \hline
        \multirow{2}{=}{Method} & \multirow{2}{=}{\centering Distance \\ Range} & 
        \multicolumn{2}{c}{Category} \\
        \cmidrule(lr){3-4}
        &  &Car  &Pedestrian \\
        \hline
        \multirow{2}{*}{M$^2$Track} & $<30m$ 
        & 68.73 / 77.14  & 44.44 / 73.33 \\
        & $\geqslant30m$  & 57.24 / 64.92 & 34.14 / 60.89 \\
        \hline
        \multirow{2}{*}{Ours} & $<30m$
        & 73.23\textcolor{mygreen}{\scriptsize $\uparrow$3.28} / 80.42\textcolor{mygreen}{\scriptsize $\uparrow$10.80}\ & 55.24\textcolor{mygreen}{\scriptsize $\uparrow$4.50} / 83.00\textcolor{mygreen}{\scriptsize $\uparrow$9.67}  \\

        &  $\geqslant30m$  & 67.62\textcolor{mygreen}{\scriptsize $\uparrow$10.38} / 73.74\textcolor{mygreen}{\scriptsize $\uparrow$8.82}&49.14\textcolor{mygreen}{\scriptsize $\uparrow$15.0} / 78.44\textcolor{mygreen}{\scriptsize $\uparrow$17.55}\\
        \hline
    \end{tabular}
    \label{tab:diff_range}
    \vspace{-2mm}
\end{table}

\noindent\textbf{The effectiveness in long-distance scenarios.} Here, we classify the trajectories into different ranges based on the initial bbox location to evaluate the tracking performance in long-distance scenarios. As shown in Table \ref{tab:diff_range}, our MVCTrack outperforms M$^2$Track, which also uses motion cues, by 10.38\%/8.82\% and 15.00\%/17.55\% in Car and Pedestrian, demonstrating the effectiveness in long-distance scenes.

\begin{table}[!h]
\vspace{-3mm}
    \caption{Comparison with different 2d images resolutions.}
    \centering
    \setlength{\tabcolsep}{3pt} 
    \renewcommand{\arraystretch}{1.2} 
    \begin{tabular}{p{0.3\linewidth}
    <{\centering}p{0.2\linewidth}<{\centering}p{0.2\linewidth}<{\centering}p{0.2\linewidth}} %
        \hline
        \multirow{2}{*}{Resolutions} &  \multicolumn{2}{c}{Category} \\
        \cmidrule(lr){2-3}
        & Car & Pedestrian \\
        \hline
        800 x 450 & 64.53 / 72.92 & 46.22 / 75.31 \\
        \hline
        1600 x 900 (origin) & \textbf{66.76} / \textbf{73.76} & \textbf{47.33} / \textbf{76.64} \\
        \hline
    \end{tabular}
    \label{tab:diff_resolutions}
    \vspace{-3mm}
\end{table}

\noindent\textbf{Robustness of 2D Segmentation Quality.} As shown in Tab~\ref{tab:diff_resolutions}, We evaluate the robustness of our method against variations in the quality of 2D segmentation results by reducing the resolution of the input images. Specifically, we decrease the resolution from the original 1600x900 to half,  simulating inaccurate segmentation results. This experiment aims to evaluate the impact of decreased segmentation precision on the generation of virtual cues and overall tracking performance. Despite the significant reduction in image resolution, our method maintains a high accuracy, achieving a tracking success and precision of 64.53/72.92 and 46.22/75.31 in car and pedestrian, outperforming similar state-of-the-art models. These results indicate that our model demonstrates resilience to lower-quality segmentation inputs, suggesting that the MVCP mechanism effectively compensates for the diminished precision by generating robust virtual cues. This highlights the adaptability of our method, even when 2D segmentation results is compromised, further emphasizing its practical applicability in real-world scenarios where segmentation quality may fluctuate.

\begin{table}[!h]
\vspace{-2mm}
    \caption{Generalization ability on M2-Track.}
    \centering
    \setlength{\tabcolsep}{3pt} 
    \renewcommand{\arraystretch}{1.2} 
    \begin{tabular}
    {p{0.18\linewidth}
    <{\centering}p{0.1\linewidth}
    <{\centering}p{0.31\linewidth}
    <{\centering}p{0.31\linewidth}<{\centering}p{0.2\linewidth}<{\centering}p{0.2\linewidth}} %
        \toprule
        \multirow{2}{=}{Strategy\\/Method} & \multirow{2}{=}{\centering Modality} & 
        \multicolumn{2}{c}{Category} \\
        \cmidrule(lr){3-4}
        &  &Car  &Pedestrian \\
        \midrule
        M$^2$Track & L+S 
        & 57.22 / 65.72  & 32.10 / 60.92 \\
        \midrule
        M$^2$Track+VC & LC+M 
        & 58.15\textcolor{mygreen}{\scriptsize $\uparrow$1.93} / 66.88\textcolor{mygreen}{\scriptsize $\uparrow$2.16}  & 36.64\textcolor{mygreen}{\scriptsize $\uparrow$4.54} / 65.28\textcolor{mygreen}{\scriptsize $\uparrow$4.36} \\
        \bottomrule
    \end{tabular}
    \label{tab:generalization}
    \vspace{-4mm}
\end{table}


\noindent\textbf{Generalization Ability on Other Trackers.} To demonstrate the generalization capability of the proposed MVCP scheme, we employ the generated virtual cues to representative M$^2$Track~\cite{m2track}. As shown in Tab. \ref{tab:generalization}, the performance of M$^2$Track improved significantly by 1.93\%/2.16\% and 4.54\%/4.36\% in Car and Pedestrian after incorporating generated virtual cues, strongly validating the effectiveness and generalization ability of our proposed MVCP mechanism. This result indicates that our MVCP method not only enhances MVCTrack but also provides substantial performance improvements for other state-of-the-art trackers. Furthermore, the plug-and-play feature of our approach enables seamless integration into existing LiDAR-based 3D SOT trackers, resulting in notable performance gains.


\noindent\textbf{Running Speed.} Our MVCTrack can achieve 32.1 FPS on single NVIDIA 3090Ti GPU and make certain improvement compared with 13.2 FPS of multi-modal MMFTrack~\cite{mmftrack}.

\noindent\textbf{Visualization results.} As shown in Fig~\ref{fig:vis}, we visualize the tracking results over the state-of-the-art method M2Track~\cite{m2track} and our baseline on the nuScenes~\cite{m2track} dataset, across diverse trajectories. In extremely sparse scenarios (rows 1 and 2), both M$^2$Track (\textcolor{red}{red bbox}) and the baseline (\textcolor{green}{green bbox}) will fail to track. In contrast, our MVCTrack (\textcolor{yellow}{yellow bbox}) along with virtual cues is able to tightly track the target. In other case (rows 3 and 4), M$^2$Track (\textcolor{red}{red bbox}) and the the baseline (\textcolor{green}{green bbox}) incorrectly track these similar objects due to insufficient points geometrics for differentiation. In comparison, our MVCTrack (\textcolor{yellow}{yellow bbox}) utilizes virtual cues to achieve robust tracking performance, closely aligning with the Ground Truth (\textcolor{blue}{blue bbox}). The consistent robustness of MVCTrack, even in challenging sparse cases, shows its effectiveness for real-world scenes.


\vspace{-1mm}
\section{CONCLUSION}
\vspace{-1mm}
In this paper, to mitigate the inherent sparsity of point cloud, we propose a Multimodal-guided Virtual Cues Projection (MVCP) mechanism aimed at enhancing the performance of 3D SOT. Based on MVCP, we construct the MVCTrack framework, which directly utilizes the generated dense 3D virtual cues alongside the raw point cloud as input to the network. The proposed MVCTrack offers three key advantages: (1) Lightweight 2D object segmentation, ensuring efficient integration; (2) Balanced point density distribution, reducing the disparity between near and far objects; and (3) A plug-in module that can seamlessly integrate with existing trackers. Extensive Experiments demonstrate that MVCTrack achieves competitive performance on the nuScenes dataset, and significantly surpass existing multi-modal 3D trackers. We hope our research can provide new insights for 3D SOT. 

\vspace{-2mm}
\bibliographystyle{IEEEtran}
\bibliography{IEEEabrv,strings,refs}

\begin{thebibliography}{10}
\providecommand{\url}[1]{#1}
\csname url@rmstyle\endcsname
\providecommand{\newblock}{\relax}
\providecommand{\bibinfo}[2]{#2}
\providecommand\BIBentrySTDinterwordspacing{\spaceskip=0pt\relax}
\providecommand\BIBentryALTinterwordstretchfactor{4}
\providecommand\BIBentryALTinterwordspacing{\spaceskip=\fontdimen2\font plus
\BIBentryALTinterwordstretchfactor\fontdimen3\font minus \fontdimen4\font\relax}
\providecommand\BIBforeignlanguage[2]{{%
\expandafter\ifx\csname l@#1\endcsname\relax
\typeout{** WARNING: IEEEtran.bst: No hyphenation pattern has been}%
\typeout{** loaded for the language `#1'. Using the pattern for}%
\typeout{** the default language instead.}%
\else
\language=\csname l@#1\endcsname
\fi
#2}}

\bibitem{probabilistic}
X.~Zhou, D.~Wang, and P.~Kr{\"a}henb{\"u}hl, ``Objects as points,'' in \emph{arXiv preprint arXiv:1904.07850}, 2019.

\bibitem{ptt}
J.~Shan, S.~Zhou, Z.~Fang, and Y.~Cui, ``Ptt: Point-track-transformer module for 3d single object tracking in point clouds,'' in \emph{2021 IEEE/RSJ International Conference on Intelligent Robots and Systems (IROS)}.\hskip 1em plus 0.5em minus 0.4em\relax IEEE, 2021, pp. 1310--1316.

\bibitem{zhou2023fastpillars}
S.~Zhou, Z.~Tian, X.~Chu, X.~Zhang, B.~Zhang, X.~Lu, C.~Feng, and Z.~Jie, ``Fastpillars: A deployment-friendly pillar-based 3d detector.''

\bibitem{eagermot}
A.~Kim, A.~Ošep, and L.~Leal-Taixé, ``Eagermot: 3d multi-object tracking via sensor fusion,'' in \emph{Proceedings of the 2021 IEEE International Conference on Robotics and Automation (ICRA)}.\hskip 1em plus 0.5em minus 0.4em\relax IEEE, 2021, pp. 11\,315--11\,321.

\bibitem{realtime}
J.~Shan \emph{et~al.}, ``Real-time 3d single object tracking with transformer,'' \emph{IEEE Transactions on Multimedia}, vol.~25, pp. 2339--2353, 2022.

\bibitem{cui2019point}
Y.~Cui, Z.~Fang, and S.~Zhou, ``Point siamese network for person tracking using 3d point clouds,'' \emph{Sensors}, vol.~20, no.~1, p. 143, 2019.

\bibitem{p2b}
H.~Qi, C.~Feng, Z.~Cao, F.~Zhao, and Y.~Xiao, ``P2b: Point-to-box network for 3d object tracking in point clouds,'' in \emph{computer vision and pattern recognition}, 2020, pp. 6329--6338.

\bibitem{3dsiamrpn}
Z.~Fang, S.~Zhou, Y.~Cui, and S.~Scherer, ``3d-siamrpn: An end-to-end learning method for real-time 3d single object tracking using raw point cloud,'' \emph{IEEE Sensors Journal}, vol.~21, no.~4, pp. 4995--5011, 2020.

\bibitem{ptt-journal}
J.~Shan, S.~Zhou, Y.~Cui, and Z.~Fang, ``Real-time 3d single object tracking with transformer,'' \emph{IEEE Transactions on Multimedia}, vol.~25, pp. 2339--2353, 2022.

\bibitem{lttr}
Y.~Cui, Z.~Fang, J.~Shan, Z.~Gu, and S.~Zhou, ``3d object tracking with transformer,'' \emph{British Machine Vision Conference}, pp. 1445--1458, 2021.

\bibitem{bat}
C.~Zheng, X.~Yan, J.~Gao, W.~Zhao, W.~Zhang, Z.~Li, and S.~Cui, ``Box-aware feature enhancement for single object tracking on point clouds,'' in \emph{Proceedings of the IEEE/CVF International Conference on Computer Vision}, 2021, pp. 13\,199--13\,208.

\bibitem{v2b}
L.~Hui, L.~Wang, M.~Cheng, J.~Xie, and J.~Yang, ``3d siamese voxel-to-bev tracker for sparse point clouds,'' \emph{Advances in Neural Information Processing Systems}, vol.~34, pp. 28\,714--28\,727, 2021.

\bibitem{siamfc}
L.~Bertinetto, J.~Valmadre, J.~F. Henriques, A.~Vedaldi, and P.~H. Torr, ``Fully-convolutional siamese networks for object tracking,'' in \emph{Computer Vision--ECCV 2016 Workshops: Amsterdam, The Netherlands, October 8-10 and 15-16, 2016, Proceedings, Part II 14}.\hskip 1em plus 0.5em minus 0.4em\relax Springer, 2016, pp. 850--865.

\bibitem{siamrpn}
B.~Li, J.~Yan, W.~Wu, Z.~Zhu, and X.~Hu, ``High performance visual tracking with siamese region proposal network,'' in \emph{Proceedings of the IEEE conference on computer vision and pattern recognition}, 2018, pp. 8971--8980.

\bibitem{siamla}
J.~Nie, Z.~He, Y.~Yang, M.~Gao, and Z.~Dong, ``Learning localization-aware target confidence for siamese visual tracking,'' \emph{IEEE Transactions on Multimedia}, 2022.

\bibitem{F-siamese}
H.~Zou, J.~Cui, X.~Kong, C.~Zhang, Y.~Liu, F.~Wen, and W.~Li, ``F-siamese tracker: A frustum-based double siamese network for 3d single object tracking,'' in \emph{IEEE/RSJ International Conference on Intelligent Robots and Systems}, 2020, pp. 8133--8139.

\bibitem{mmftrack}
Z.~Li \emph{et~al.}, ``Mmf-track: Multi-modal multi-level fusion for 3d single object tracking,'' \emph{IEEE Transactions on Intelligent Vehicles}, 2023.

\bibitem{mvp}
T.~Yin, X.~Zhou, and P.~Kr{\"a}henb{\"u}hl, ``Multimodal virtual point 3d detection,'' in \emph{Advances in Neural Information Processing Systems}, vol.~34, 2021, pp. 16\,494--16\,507.

\bibitem{yolov6}
C.~Li \emph{et~al.}, ``Yolov6: A single-stage object detection framework for industrial applications,'' \emph{arXiv preprint arXiv:2209.02976}, 2022.

\bibitem{yolov7}
C.-Y. Wang, A.~Bochkovskiy, and H.-Y.~M. Liao, ``Yolov7: Trainable bag-of-freebies sets new state-of-the-art for real-time object detectors,'' in \emph{Proceedings of the IEEE/CVF Conference on Computer Vision and Pattern Recognition}, 2023, pp. 7464--7475.

\bibitem{nuScenes}
H.~Caesar, V.~Bankiti, A.~H. Lang, S.~Vora, V.~E. Liong, Q.~Xu, A.~Krishnan, Y.~Pan, G.~Baldan, and O.~Beijbom, ``nuscenes: A multimodal dataset for autonomous driving,'' in \emph{Proceedings of the IEEE/CVF conference on computer vision and pattern recognition}, 2020, pp. 11\,621--11\,631.

\bibitem{p2p}
J.~Nie, F.~Xie, S.~Zhou, X.~Zhou, D.-K. Chae, and Z.~He, ``P2p: Part-to-part motion cues guide a strong tracking framework for lidar point clouds,'' \emph{arXiv e-prints}, pp. arXiv--2407, 2024.

\bibitem{sc3d}
S.~Giancola, J.~Zarzar, B.~Ghanem, S.~Giancola, and J.~Zarzar, ``Leveraging shape completion for 3d siamese tracking,'' in \emph{Proceedings of the IEEE/CVF conference on computer vision and pattern recognition}, 2019, pp. 1359--1368.

\bibitem{votenet}
C.~R. Qi, O.~Litany, K.~He, and L.~J. Guibas, ``Deep hough voting for 3d object detection in point clouds,'' in \emph{proceedings of the IEEE/CVF International Conference on Computer Vision}, 2019, pp. 9277--9286.

\bibitem{stnet}
L.~Hui, L.~Wang, L.~Tang, K.~Lan, J.~Xie, and J.~Yang, ``3d siamese transformer network for single object tracking on point clouds,'' in \emph{Computer Vision--ECCV 2022: 17th European Conference, Tel Aviv, Israel, October 23--27, 2022, Proceedings, Part II}.\hskip 1em plus 0.5em minus 0.4em\relax Springer, 2022, pp. 293--310.

\bibitem{glt}
J.~Nie, Z.~He, Y.~Yang, M.~Gao, and J.~Zhang, ``Glt-t: Global-local transformer voting for 3d single object tracking in point clouds,'' in \emph{Proceedings of the AAAI Conference on Artificial Intelligence}, 2023, pp. 1957--1965.

\bibitem{pttr}
C.~Zhou, Z.~Luo, Y.~Luo, T.~Liu, L.~Pan, Z.~Cai, H.~Zhao, and S.~Lu, ``Pttr: Relational 3d point cloud object tracking with transformer,'' in \emph{Proceedings of the IEEE/CVF Conference on Computer Vision and Pattern Recognition}, 2022, pp. 8531--8540.

\bibitem{cmt}
Z.~Guo, Y.~Mao, W.~Zhou, M.~Wang, and H.~Li, ``Cmt: Context-matching-guided transformer for 3d tracking in point clouds,'' in \emph{Computer Vision--ECCV 2022: 17th European Conference, Tel Aviv, Israel, October 23--27, 2022, Proceedings, Part XXII}.\hskip 1em plus 0.5em minus 0.4em\relax Springer, 2022, pp. 95--111.

\bibitem{dmt}
Y.~Xia, Q.~Wu, W.~Li, A.~B. Chan, and U.~Stilla, ``A lightweight and detector-free 3d single object tracker on point clouds,'' \emph{IEEE Transactions on Intelligent Transportation Systems}, 2023.

\bibitem{osp2b}
J.~Nie, Z.~He, Y.~Yang, Z.~Bao, M.~Gao, and J.~Zhang, ``Osp2b: One-stage point-to-box network for 3d siamese tracking,'' in \emph{Proceedings of the Thirty-Second International Joint Conference on Artificial Intelligence}, 2023, pp. 1285--1293.

\bibitem{tat}
K.~Lan, H.~Jiang, and J.~Xie, ``Temporal-aware siamese tracker: Integrate temporal context for 3d object tracking,'' in \emph{Proceedings of the Asian Conference on Computer Vision}, 2022, pp. 399--414.

\bibitem{synctrack}
T.~Ma, M.~Wang, J.~Xiao, H.~Wu, and Y.~Liu, ``Synchronize feature extracting and matching: A single branch framework for 3d object tracking,'' in \emph{Proceedings of the IEEE/CVF International Conference on Computer Vision}, 2023, pp. 9953--9963.

\bibitem{xu2024pillartrack}
W.~Xu, S.~Zhou, and Z.~Yuan, ``Pillartrack: Redesigning pillar-based transformer network for single object tracking on point clouds,'' \emph{arXiv preprint arXiv:2404.07495}, 2024.

\bibitem{m2track}
C.~Zheng, X.~Yan, H.~Zhang, B.~Wang, S.~Cheng, S.~Cui, and Z.~Li, ``Beyond 3d siamese tracking: A motion-centric paradigm for 3d single object tracking in point clouds,'' in \emph{Proceedings of the IEEE/CVF Conference on Computer Vision and Pattern Recognition}, 2022, pp. 8111--8120.

\bibitem{m2track++}
C.~Zheng, X.~Yan, H.~Zhang, S.~Cui, and Z.~Li, ``An effective motion-centric paradigm for 3d single object tracking in point clouds,'' \emph{IEEE Transactions on Pattern Analysis and Machine Intelligence}, 2023.

\bibitem{smat}
Y.~Cui, J.~Shan, Z.~Gu, Z.~Li, and Z.~Fang, ``Exploiting more information in sparse point cloud for 3d single object tracking,'' \emph{IEEE Robotics and Automation Letters}, vol.~7, no.~4, pp. 11\,926--11\,933, 2022.

\bibitem{stttracker}
Y.~Cui, Z.~Li, and Z.~Fang, ``Sttracker: Spatio-temporal tracker for 3d single object tracking,'' \emph{IEEE Robotics and Automation Letters}, 2023.

\bibitem{cutrack}
J.~Nie, Z.~He, X.~Lv, X.~Zhou, D.-K. Chae, and F.~Xie, ``Towards category unification of 3d single object tracking on point clouds,'' in \emph{The Twelfth International Conference on Learning Representations}, 2024.

\bibitem{flowtrck}
S.~Li, Y.~Cui, L.~Zhiheng, and Z.~Fang, ``Flowtrack: Point-level flow network for 3d single object tracking,'' \emph{arXiv preprint arXiv:2407.01959}, 2024.

\bibitem{mbptrack}
T.-X. Xu, Y.-C. Guo, Y.-K. Lai, and S.-H. Zhang, ``Mbptrack: Improving 3d point cloud tracking with memory networks and box priors,'' \emph{arXiv preprint arXiv:2303.05071}, 2023.

\bibitem{otb2013}
Y.~Wu, J.~Lim, and M.-H. Yang, ``Online object tracking: A benchmark,'' in \emph{Proceedings of the IEEE conference on computer vision and pattern recognition}, 2013, pp. 2411--2418.

\bibitem{kristan2016novel}
M.~Kristan, J.~Matas, A.~Leonardis, T.~Voj{\'\i}{\v{r}}, R.~Pflugfelder, G.~Fernandez, G.~Nebehay, F.~Porikli, and L.~{\v{C}}ehovin, ``A novel performance evaluation methodology for single-target trackers,'' \emph{IEEE transactions on pattern analysis and machine intelligence}, vol.~38, no.~11, pp. 2137--2155, 2016.

\end{thebibliography}

\end{document}